\documentclass[letterpaper, 10 pt, conference]{ieeeconf}  

\IEEEoverridecommandlockouts                              

\usepackage{graphics} 
\usepackage{epsfig} 
\usepackage{mathptmx} 
\usepackage{times} 
\usepackage{amsmath} 
\usepackage{amssymb}  
\usepackage{booktabs}
\usepackage{multirow}
\usepackage{balance}
\usepackage{rotating}
\usepackage{url}

\title{\LARGE \bf
Superpoint-guided Semi-supervised Semantic Segmentation of 3D Point Clouds
}
\author{Shuang Deng, Qiulei Dong$^{*}$, Bo Liu, and Zhanyi Hu
\thanks{This work was supported by the National Natural Science Foundation of China (Grant Nos. U1805264 and 61991423), the Strategic Priority Research Program of the Chinese Academy of Sciences (XDB32050100), and the Open Research Fund from Key Laboratory of Intelligent Infrared Perception, Chinese Academy of Sciences.}
\thanks{S. Deng, Q. Dong, and B. Liu are with the National Laboratory of Pattern Recognition, Institute of Automation, Chinese Academy of Sciences, Beijing 100190, China, and also with the School of Artificial Intelligence, University of Chinese Academy of Sciences, Beijing 100049, China, and also with the Center for Excellence in Brain Science and Intelligence Technology, Chinese Academy of Sciences, Beijing 100190, China (e-mail: [shuang.deng, qldong]@nlpr.ia.ac.cn, liubo2017@ia.ac.cn).}
\thanks{Z. Hu is with the National Laboratory of Pattern Recognition, Institute of Automation, Chinese Academy of Sciences, Beijing 100190, China, and also with the School of Artificial Intelligence, University of Chinese Academy of Sciences, Beijing 100049, China (e-mail: huzy@nlpr.ia.ac.cn).}
\thanks{*Corresponding author.}
}

\begin{document}

\maketitle
\thispagestyle{empty}
\pagestyle{empty}

\begin{abstract}
3D point cloud semantic segmentation is a challenging topic in the computer vision field. Most of the existing methods in literature require a large amount of fully labeled training data, but it is extremely time-consuming to obtain these training data by manually labeling massive point clouds. Addressing this problem, we propose a superpoint-guided semi-supervised segmentation network for 3D point clouds, which jointly utilizes a small portion of labeled scene point clouds and a large number of unlabeled point clouds for network training. The proposed network is iteratively updated with its predicted pseudo labels, where a superpoint generation module is introduced for extracting superpoints from 3D point clouds, and a pseudo-label optimization module is explored for automatically assigning pseudo labels to the unlabeled points under the constraint of the extracted superpoints. Additionally, there are some 3D points without pseudo-label supervision. We propose an edge prediction module to constrain features of edge points. A superpoint feature aggregation module and a superpoint feature consistency loss function are introduced to smooth superpoint features. Extensive experimental results on two 3D public datasets demonstrate that our method can achieve better performance than several state-of-the-art point cloud segmentation networks and several popular semi-supervised segmentation methods with few labeled scenes.
\end{abstract}

\begin{figure*}
	\centering
	\includegraphics[width=2 \columnwidth]{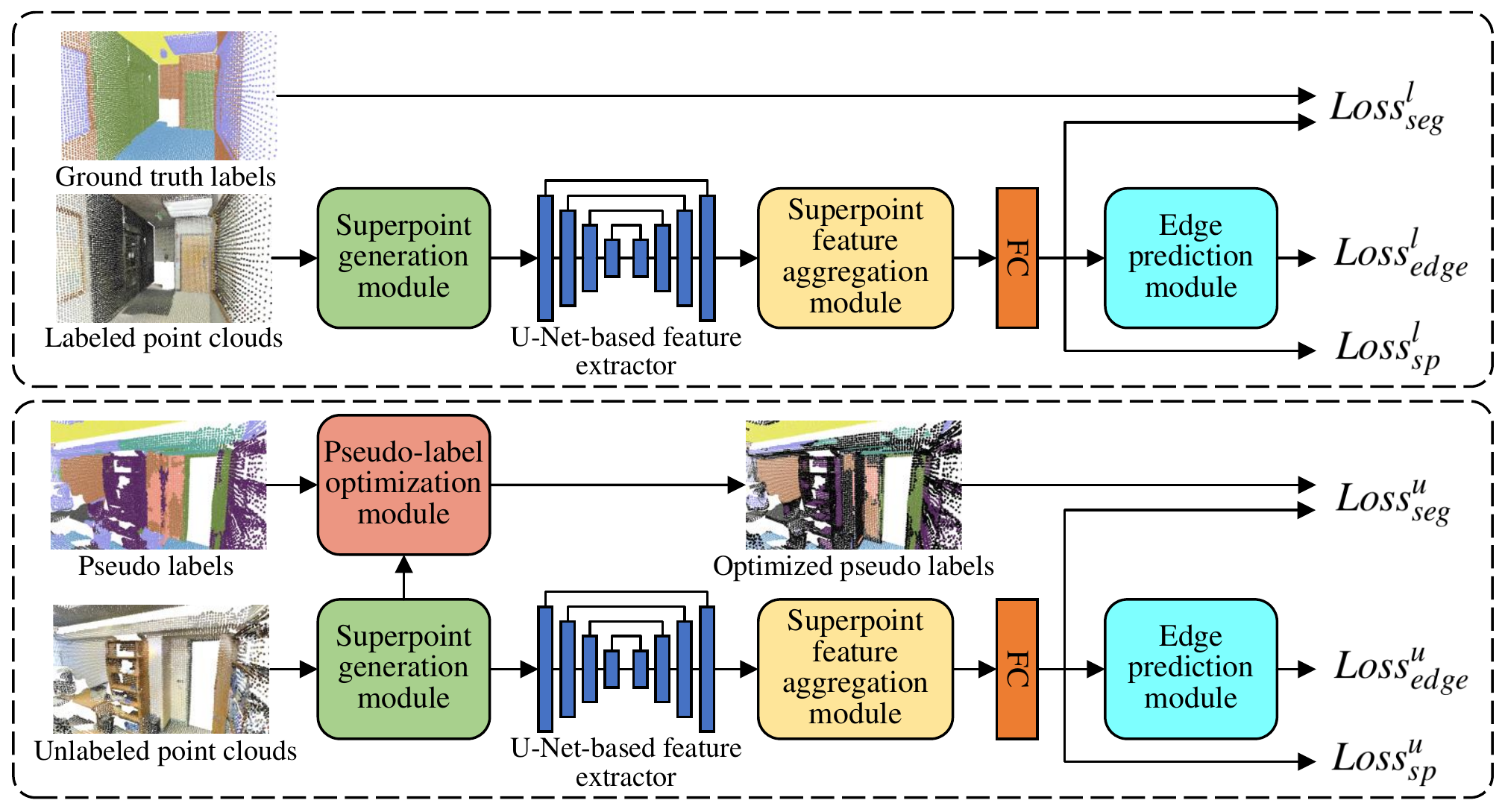}
	\caption{Architecture of the proposed network. The structures in two dashed boxes are for labeled and unlabeled point clouds respectively. 
		FC represents the fully-connected layer.
	}
	\label{architecture}
\end{figure*}

\section{Introduction}
3D point cloud semantic segmentation draws increasing attention in the field of computer vision. In recent years, a large number of Deep Neural Networks (DNNs)~\cite{2017pointnet, 2017pointnet++, 2018sscn, 2018pointcnn, 2018spgraph, 2018dgcnn, 2019pointweb, 2019sspspg, 2020randlanet, 2020grid, 2021rtn, 2021scfnet, 2021ganet} for point cloud semantic segmentation have been proposed. Although these methods have a great ability to obtain the semantic features of point clouds, most of them require a large number of accurately labeled 3D scenes, and manually labeling point clouds is time and labor-intensive.

Recently, some weakly supervised segmentation methods~\cite{2012active, 2020wsspcs, 2020ss3ld, 2021spcs, 2021sspcnet} for 3D point clouds have been proposed,
which could be roughly divided into two groups according to two different kinds of training datasets: (1) weakly supervised methods whose training dataset contains a small portion of labeled points sampled from each 3D training scene; (2) weakly supervised methods (also called semi-supervised methods) whose training dataset contains a small portion of labeled 3D scenes.  
The former group of methods~\cite{2020wsspcs, 2021sspcnet} require point sampling for all 3D scenes, and the point clouds sampled from some dense 3D scenes will still be somewhat dense, and the labor costs of assigning point labels will not be reduced too much. 
Compared with the former group of methods, the semi-supervised methods~\cite{2012active, 2020ss3ld, 2021spcs} are able to significantly reduce labeling costs, hence, we focus on the semi-supervised point cloud segmentation probelm in this paper.

For solving the semi-supervised semantic segmentation problem for 3D point clouds, the two methods~\cite{2012active, 2020ss3ld} introduce additional information of point clouds. Wang et al.~\cite{2012active} utilizes expert knowledge and Mei et al.~\cite{2020ss3ld} considers the consistency of scans stream. Besides, the point clouds used by the methods~\cite{2012active, 2021spcs} are CAD models, which are much simpler than 3D scenes. All these three methods~\cite{2012active, 2020ss3ld, 2021spcs} do not consider the prior geometry and color knowledge of point clouds, which is useful for pseudo-label selecting.
In addition, there are some methods~\cite{2017pimodel, 2017meanteacher, 2017ssgan, 2021dtc} to solve the semi-supervised segmentation problem for 2D images. However, since 3D point cloud is an unordered and irregular structure, these methods cannot be applied to 3D point clouds directly.

It is noted that several existing works~\cite{2018spgraph, 2019sspspg, 2020pointnl, 2021sspcnet} for 3D point cloud semantic segmentation utilize superpoints to improve their performances.
A few methods~\cite{2018spgraph, 2020pointnl, 2021sspcnet} geometrically partition the point clouds by minimizing a global energy function. These methods do not consider the color information of 3D point clouds, where some classes of objects are only different in color from the surrounding objects (\emph{i.e.} window and board). And minimizing the global energy function is time-consuming. Landrieu et al.~\cite{2019sspspg} formulates superpoints generation as a deep metric learning problem. But this partition method requires semantic information of the 3D point clouds.

Addressing the aforementioned issues, we propose a superpoint-guided semi-supervised segmentation network for 3D point clouds. The labeled and unlabeled point clouds will be processed in different ways. We use the ground truth labels to supervise the labeled point clouds. And the pseudo labels predicted from unlabeled point clouds are used for self-training. Since the pseudo labels are not completely accurate, we utilize the superpoints to optimize pseudo labels. Specifically, we propose a superpoint generation module, named as SPG module, to combine the superpoints produced by geometry-based and color-based Region Growing algorithms~\cite{1994regiongrowin}, and a pseudo-label optimization module, named as PLO module, to modify and delete pseudo labels with low confidence in each superpoint. 
There are some 3D points without pseudo-label supervision. We propose an edge prediction module, named as EP module, to constrain the features from edge points of geometry and color. A superpoint feature aggregation module, named as SPFA module, and a superpoint feature consistency loss function are introduced to smooth the point features in each superpoint. 

In sum, the main contributions of this paper include:
\begin{itemize}
	\item For solving the semi-supervised semantic segmentation problem of 3D point clouds effectively and efficiently, we utilize the superpoints generated by combining geometry-based and color-based Region Growing algorithms to optimize pseudo labels predicted from unlabeled point clouds.
	\item We propose an edge prediction module, a superpoint feature aggregation module and a superpoint feature consistency loss function for constraining point features without pseudo labels. 
	\item We propose the superpoint-guided semi-supervised segmentation network for 3D point clouds. The experimental results on two 3D public datasets show that the proposed method outperforms several state-of-the-art point cloud segmentation networks and several popular semi-supervised segmentation methods with few labeled scenes.
\end{itemize}

\section{Superpoint-guided Semi-supervised Segmentation Network}
In this section, we propose the superpoint-guided semi-supervised segmentation network for 3D point clouds. Firstly, we introduce the architecture of the proposed network. Secondly, we describe the details of the superpoint generation module (SPG module), the pseudo-label optimization module (PLO module), the edge prediction module (EP module), the superpoint feature aggregation module (SPFA module) and the superpoint feature consistency loss function respectively. Lastly, we end up with the final training loss of the network.

\subsection{Architecture}
As shown in Fig. \ref{architecture}, our end-to-end superpoint-guided semi-supervised segmentation network consists of two branches. The inputs of one branch are labeled point clouds and their ground truth labels, and the other branch are unlabeled point clouds and their pseudo labels. The pseudo labels are predicted by our network from unlabeled point clouds. Both branches consist of a superpoint generation module (SPG module), an feature extractor based on U-Net~\cite{2015unet}, a superpoint feature aggregation module (SPFA module), a fully connected layer (FC), and an edge prediction module (EP module). And their parameters are shared.
For the branch of unlabeled point clouds, there is a pseudo-label optimization module (PLO module) to optimize the pseudo labels.
The U-Net-based feature extractor consists of four encoder layers and four decoder layers. The encoder layers are Local Feature Aggregation layers in RandLA-Net~\cite{2020randlanet}, and the decoder layers are MLPs.

\begin{figure}
	\centering
	\includegraphics[width=1 \columnwidth]{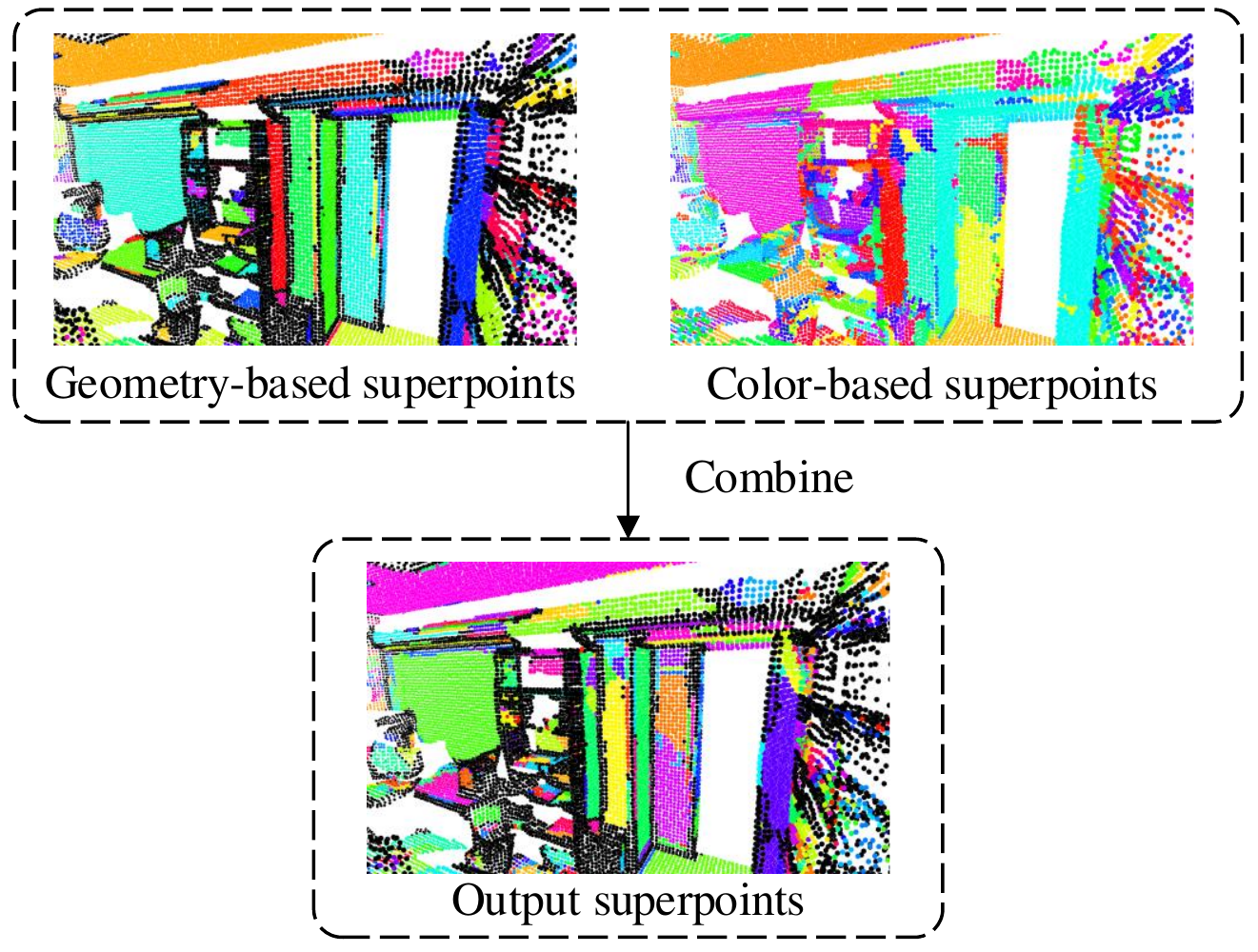}
	\caption{The process of combining superpoints produced by geometry-based and color-based Region Growing algorithms. The black points are not clustered as superpoints due to the curvature threshold in the geometry-based Growing Region algorithm.}
	\label{spg}
\end{figure}

\begin{figure}
	\centering
	\includegraphics[width=1 \columnwidth]{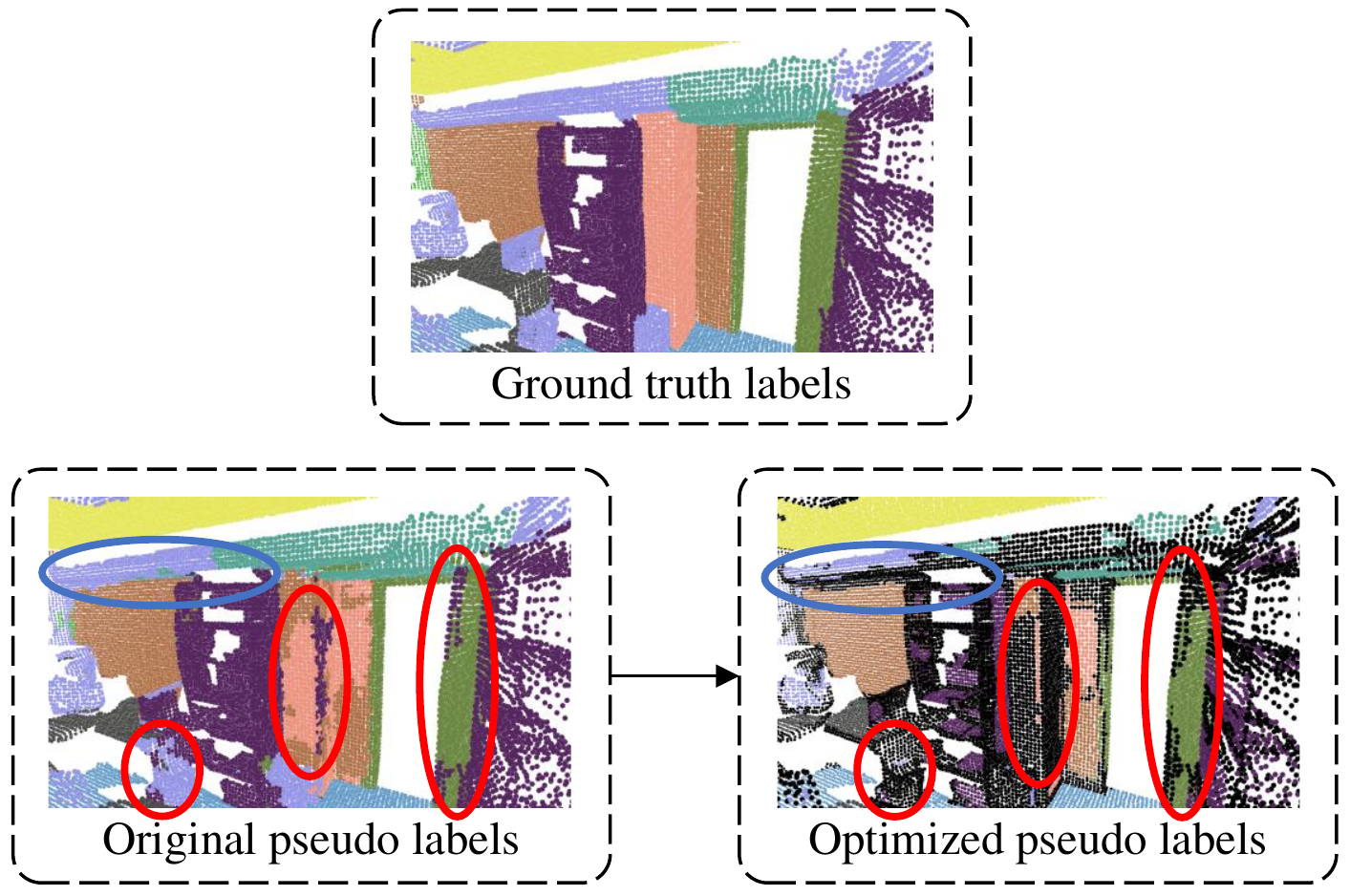}
	\caption{The process of optimizing pseudo labels. Some pseudo labels inside the red and blue circles are incorrect. The black points have no pseudo labels after optimizing. }
	\label{plo}
\end{figure}

When a labeled 3D point cloud 
$\mathbf{P}^l = \{\boldsymbol{p}^l_1, \boldsymbol{p}^l_2, ..., \boldsymbol{p}^l_{N^l}\} \in \mathbb{R}^{N^l \times 6}$
and its one-hot ground truth labels 
$\mathbf{Y}^l = \{\boldsymbol{y}^l_1, \boldsymbol{y}^l_2, ..., \boldsymbol{y}^l_{N^l}\} \in \mathbb{R}^{N^l \times C}$,
and an unlabeled point cloud 
$\mathbf{P}^u = \{\boldsymbol{p}^u_1, \boldsymbol{p}^u_2, ..., \boldsymbol{p}^u_{N^u}\} \in \mathbb{R}^{N^u \times 6}$
and its one-hot pseudo labels
$\mathbf{Y}^u = \{\boldsymbol{y}^u_1, \boldsymbol{y}^u_2, ..., \boldsymbol{y}^u_{N^u}\} \in \mathbb{R}^{N^u \times C}$
are given, where $N^l$ and $N^u$ are the number of points, $6$ denotes the XYZ dimensions and RGB dimensions, and $C$ is the number of semantic classes, we firstly send $\mathbf{P}^l$ and $\mathbf{P}^u$ to the SPG module to generate their superpoint collections
$\mathbf{S}^l = \{\mathbf{S}^l_1, \mathbf{S}^l_2, ..., \mathbf{S}^l_{M^l}\}$
and
$\mathbf{S}^u = \{\mathbf{S}^u_1, \mathbf{S}^u_2, ..., \mathbf{S}^u_{M^u}\}$,
where $M^l$ and $M^u$ are the number of superpoints.
For the $i^{th}$ superpoint in $\mathbf{S}^l$,
$\mathbf{S}^l_i = \{\boldsymbol{p}^l_{i_1}, \boldsymbol{p}^l_{i_2}, ..., \boldsymbol{p}^l_{i_n}\} \in \mathbb{R}^{n \times 6}$,
where $n$ is the number of points in this superpoint, similar in $\mathbf{S}^u$.
Secondly, we send $\mathbf{P}^l$ and $\mathbf{P}^u$ to the U-Net-based feature extractor to construct their high-level representations 
$\mathbf{F}^l = \{\boldsymbol{f}^l_1, \boldsymbol{f}^l_2, ..., \boldsymbol{f}^l_{N^l}\} \in \mathbb{R}^{N^l \times C_h}$
and
$\mathbf{F}^u = \{\boldsymbol{f}^u_1, \boldsymbol{f}^u_2, ..., \boldsymbol{f}^u_{N^u}\} \in \mathbb{R}^{N^u \times C_h}$,
where $C_h$ is the dimension of high-level features.
Then we send $\mathbf{F}^l$ and $\mathbf{F}^u$ to the SPFA module to get feature maps
$\mathbf{G}^l = \{\boldsymbol{g}^l_1, \boldsymbol{g}^l_2, ..., \boldsymbol{g}^l_{N^l}\} \in \mathbb{R}^{N^l \times C_h}$
and
$\mathbf{G}^u = \{\boldsymbol{g}^u_1, \boldsymbol{g}^u_2, ..., \boldsymbol{g}^u_{N^u}\} \in \mathbb{R}^{N^u \times C_h}$
for feature smoothing in superpoints. 
After a FC layer, we obtain the final feature maps 
$\mathbf{X}^l = \{\boldsymbol{x}^l_1, \boldsymbol{x}^l_2, ..., \boldsymbol{x}^l_{N^l}\} \in \mathbb{R}^{N^l \times C}$
and
$\mathbf{X}^u = \{\boldsymbol{x}^u_1, \boldsymbol{x}^u_2, ..., \boldsymbol{x}^u_{N^u}\} \in \mathbb{R}^{N^u \times C}$.

\subsection{Superpoint Generation Module} 
We propose a novel superpoint generation module, named as SPG, to produce superpoints effectively and efficiently.
The geometric and color characteristics of classes of objects in scenes are different.
Some classes of objects are different in geometry and color from the surrounding objects (\emph{i.e.} chair and table). But there are also some classes of objects are only different in geometry from the surrounding objects (\emph{i.e.} beam and column), or only different in color from the surrounding objects (\emph{i.e.} window and board).
Some existing superpoint generation methods~\cite{2018spgraph, 2020pointnl, 2021sspcnet} can only geometrically partition the 3D point clouds, which ignore the color information. 
The proposed SPG module combines geometry-based and color-based superpoints produced by the Region Growing algorithm~\cite{1994regiongrowin}, which has low computational complexity.

In each iteration of the geometry-based Region Growing algorithm, there is a point with minimum curvature value to be selected as a seed and added to a seeds set and a superpoint. Then, the following three steps are repeated until there are no point in the seeds set: 
(1) Finding the neighbouring points of seeds and testing their angles between their normals, these neighbouring points will be added to the current superpoint if the angles are less than the threshold value $t_{ang}$.
(2) If the curvatures of these neighbouring points are less than the threshold value $t_{cvt}$, then these points are added to the seeds set. 
(3) Current seeds are removed from the seeds set. 
If there are no unsegmented points whose curvatures are smaller than the threshold value $t_{cvt}$ in the scene, the process of iterations is terminated. Due to the curvature threshold $t_{cvt}$, some points will not be clustered to superpoints.

The color-based Region Growing algorithm is similar to the geometry-based ones. There are two main differences in the color-based algorithm. The first one is that it uses color instead of normals. The color threshold value is $t_{clr}$. The second one is that it uses the merging algorithm for segmentation control. Two neighbouring clusters with a small difference between average color are merged together. The color-based Region Growing algorithm has no curvature threshold, so every point can be clustered.

After obtaining the superpoints from the geometry-based and color-based Region Growing algorithms, we over-segment every geometry-based superpoint based on the color-based superpoints, which can be seen in Fig. \ref{spg}. It is noted that the geometric edge points will not be clustered as superpoints due to the curvature threshold $t_{cvt}$. The final merged superpoints $\mathbf{S}^l$ and $\mathbf{S}^u$ could be used by the PLO module, the SPFA module, and the superpoint feature consistency loss function.

\subsection{Pseudo-label Optimization Module}
Since the pseudo labels $\mathbf{Y}^u$ predicted by the network are not completely accurate, and the points in same superpoint should have same semantic labels in most cases, we utilize the superpoints to optimize pseudo labels. We propose a novel pseudo-label optimization module, named as PLO module, to modify and delete pseudo labels with low confidence. 

As shown from the red circle areas in the second row of Fig. \ref{plo}, incorrect pseudo labels generally have no geometric and color rules. So we can constrain pseudo labels by the geometry and color-based superpoints. Specifically, for a superpoint $\mathbf{S}^u_i (i = 1, 2, ..., M^u)$ with $n$ points, we first count the number of points contained in each semantic category $n_j (\sum_{j=1}^C n_j = n)$. Then we find the category $c_i$ that contains the most points, which can be formulated as: 
\begin{equation}
	\begin{split}
		&c_i = \mathop{\arg\max}_{j} \ \ (n_j). \\
	\end{split}
	\label{equ_plo}
\end{equation}
If $n_{c_i} > t_{plo} \times n$, where $t_{plo}$ is a ratio parameter, we modify all the pseudo labels in superpoint $\mathbf{S}^l_i$ to $c_i$, otherwise all the pseudo labels in this superpoint will be deleted. We also delete the pseudo labels of points which are not clustered as superpoints in the geometry-based Growing Region algorithm~\cite{1994regiongrowin}. After above operations being done on all superpoints in the unlabeled point clouds, the optimized pseudo labels $\mathbf{\bar{Y}}^u = \{\boldsymbol{\bar{y}}^u_1, \boldsymbol{\bar{y}}^u_2, ..., \boldsymbol{\bar{y}}^u_{N^u}\} \in \mathbb{R}^{N^u \times C}$ are shown in the second row of Fig. \ref{plo}.

\subsection{Edge Prediction Module}
The geometry-based Region Growing algorithm~\cite{1994regiongrowin} does not contain edge points due to the curvature threshold setting. And the predicted pseudo labels of geometric edge points are usually unstable, which can be seen from the area inside the blue circle in the second row of Fig. \ref{plo}. So we design an edge prediction module, named as EP module, to constrain the features of edge points.
We consider not only geometric edge points, but also color edge points. The geometric edge points are composed of points that are not clustered by the geometry-based region growing algorithm. The color edge points are those points whose neighboring points do not belong to the same color-based superpoint. 

The EP module consists of two FC layers, which reduce the number of feature channels to two. The activation function of the first FC layer is Leaky ReLU (LReLU)~\cite{2013lrelu}. The activation function of the second FC layer is Sigmoid. For the features of unlabeled point cloud $\mathbf{X}^u$, the outputs of the EP module are 
$\mathbf{E}^u = \{\boldsymbol{e}^u_1, \boldsymbol{e}^u_2, ..., \boldsymbol{e}^u_{N^u}\} \in \mathbb{R}^{N^u \times 2}$,
which can be formulated as:
\begin{equation}
	\begin{split}
		&\boldsymbol{e}^u_i = \textup{Sigmoid}(\textup{FC}(\textup{LReLU}(\textup{FC}(\boldsymbol{x}^u_i)))) \\
	\end{split}
	\label{equ_ep}
\end{equation}
where $\boldsymbol{e}^u_i $ is the $i$-th element of $\mathbf{E}^u$. The labels of EP module for the unlabeled point cloud $\mathbf{P}^u$ are 
$\mathbf{\hat{E}}^u = \{\boldsymbol{\hat{e}}^u_1, \boldsymbol{\hat{e}}^u_2, ..., \boldsymbol{\hat{e}}^u_{N^u}\} \in \mathbb{R}^{N^u \times 2}$, where the values of edge points are $1$, otherwise $0$. So the edge prediction loss function for the unlabeled point cloud  $Loss^u_{edge}$ is:
\begin{equation}
	\begin{split}
		&Loss^u_{edge} = \sum_{i=1}^{N^u} \sum_{c=1}^2 -\boldsymbol{\hat{e}}^u_{i,c}\textup{log}(\boldsymbol{e}^u_{i,c}) - (1 - \boldsymbol{\hat{e}}^u_{i,c})\textup{log}(1 - \boldsymbol{e}^u_{i,c}) \\
	\end{split}
	\label{equ_loss_edge}
\end{equation}
where $\boldsymbol{e}^u_{i,c}$ is the $c$-th channel of $\boldsymbol{e}^u_i$. The edge prediction loss function for the labeled point cloud $Loss^l_{edge}$ is obtained by the same way.

\subsection{Smoothing Superpoint Features}
In the PLO module, the pseudo labels of some superpoints are deleted, the features in these superpoints are not constrained. Besides, the points within same superpoint should have similar semantic features in most cases. So we propose a superpoint feature aggregation module, named as SPFA module, and a superpoint feature consistency loss function to smooth superpoint features.

We first introduce the SPFA module. For the $i$-th clustered point in the unlabeled point cloud $\boldsymbol{p}^u_i$, we randomly sample $K$ points
$\boldsymbol{p}^u_{i_1}, \boldsymbol{p}^u_{i_2}, ..., \boldsymbol{p}^u_{i_K}$
within the same superpoint as $\boldsymbol{p}^u_i$, and thier high-level features
$\boldsymbol{f}^u_{i_1}, \boldsymbol{f}^u_{i_2}, ..., \boldsymbol{f}^u_{i_K}$. The aggregated feature $\boldsymbol{g}^u_i$ for the point $\boldsymbol{p}^u_i$ is obtained by:
\begin{equation}
	\begin{split}
		&\boldsymbol{g}^u_i = \frac{(\boldsymbol{f}^u_i + \sum_{k=1}^{K} \boldsymbol{f}^u_{i_k})}{2}. \\
	\end{split}
	\label{equ_spfa}
\end{equation}
Obtaining $\boldsymbol{g}^l_i$ is in the same way.

Then we introduce the superpoint feature consistency loss functions $Loss^l_{sp}$ and $Loss^u_{sp}$. We use the variance function as the metric criterion of smoothness. For the features of unlabeled point cloud $\mathbf{X}^u$, the loss function $Loss^u_{sp}$ is formulated as:
\begin{equation}
	\begin{split}
		&Loss^u_{sp} = \sum_{i=1}^{N^u} \sum_{c=1}^{C} w^u_i(\boldsymbol{x}^u_{i,c} - \frac{\sum_{k=1}^{K} \boldsymbol{x}^u_{i_k,c}}{K})^2\\
	\end{split}
	\label{equ_loss_sp}
\end{equation}
where $w^u_i$ is a boolean value whether $\boldsymbol{p}^u_i$ is within a superpoint. $Loss^l_{sp}$ is obtained in the same way.

\subsection{Training Loss}
We introduce the final training loss of the superpoint-guided semi-supervised segmentation network.
For the labeled point clouds, we calculate a multi-class cross-entropy loss $Loss^l_{seg}$ between $\mathbf{Y}^l$ and the Softmax of features $\mathbf{X}^l$ as follows:
\begin{equation}
	\begin{split}
		&Loss^l_{seg} = -\sum_{i=1}^{N^l} \sum_{c=1}^{C} \boldsymbol{y}^l_{i,c}\textup{log}(\textup{Softmax}(\boldsymbol{x}^l_{i,c}))\\
	\end{split}
	\label{equ_loss_seg_l}
\end{equation}
where $\boldsymbol{y}^l_{i,c}$ is the $c$-th channel of $\boldsymbol{y}^l_i$.
For the unlabeled point clouds, we calculate a weighted multi-class cross-entropy loss $Loss^u_{seg}$ between $\mathbf{\bar{Y}}^u$ and features $\mathbf{X}^u$ as follows:
\begin{equation}
	\begin{split}
		&Loss^u_{seg} = -\sum_{i=1}^{N^u} \sum_{c=1}^{C} \bar{w}^u_i\boldsymbol{\bar{y}}^u_{i,c}\textup{log}(\textup{Softmax}(\boldsymbol{x}^u_{i,c}))\\
	\end{split}
	\label{equ_loss_seg_u}
\end{equation}
where $\bar{w}^u_i$ is a boolean value whether $\boldsymbol{p}^u_i$ has an optimized pseudo label after PLO module. 
The final loss function is formulated as:
\begin{equation}
	\begin{split}
		Loss = &Loss^l_{seg} + Loss^u_{seg} + Loss^l_{edge} + Loss^u_{edge} + \\
		&Loss^l_{sp} + Loss^u_{sp}. \\
	\end{split}
	\label{equ_loss_all}
\end{equation}


\begin{table}[t]
	\begin{center}
		\caption{Semantic segmentation results (\%) on the S3DIS dataset (Area-5).} \label{tab_s3dis}
		\resizebox{0.95\columnwidth}{!}{
			\begin{tabular}{|c|c|c|c|c|}
				\hline
				\rule{0pt}{8pt}&Methods	&mIoU	&mAcc	&OA\\
				\hline
				\multirow{7}{*}{\begin{sideways}20\%\end{sideways}}
				\rule{0pt}{8pt}&RandLA-Net~\cite{2020randlanet}	&50.90	&60.76	&81.24\\
				\rule{0pt}{8pt}&GA-Net~\cite{2021ganet}	&52.12	&61.76 	&81.46\\
				\rule{0pt}{8pt}&SCF-Net~\cite{2021scfnet}	&51.78	&61.19	&81.61\\			
				\cline{2-5}
				\rule{0pt}{8pt}&$\pi$-Model~\cite{2017pimodel}	&51.58	&59.46	&82.09\\
				\rule{0pt}{8pt}&Mean Teacher~\cite{2017meanteacher}	&51.44	&62.27	&81.70\\
				\rule{0pt}{8pt}&Pseudo-Labels~\cite{2013pseudolabels}	&52.21	&63.76	&82.39\\
				\cline{2-5}
				\rule{0pt}{8pt}&Ours	&\textbf{55.49}	&\textbf{65.45}	&\textbf{83.55}\\
				\hline
				\multirow{7}{*}{\begin{sideways}10\%\end{sideways}}
				\rule{0pt}{8pt}&RandLA-Net~\cite{2020randlanet}	&45.64	&58.58	&79.08\\
				\rule{0pt}{8pt}&GA-Net~\cite{2021ganet}	&43.85	&52.60	&78.41\\
				\rule{0pt}{8pt}&SCF-Net~\cite{2021scfnet}	&42.64	&53.16	&76.54\\			
				\cline{2-5}
				\rule{0pt}{8pt}&$\pi$-Model~\cite{2017pimodel}	&46.05	&57.49	&80.26\\
				\rule{0pt}{8pt}&Mean Teacher~\cite{2017meanteacher}	&46.72	&57.84	&80.50\\
				\rule{0pt}{8pt}&Pseudo-Labels~\cite{2013pseudolabels}	&47.78	&61.40	&81.13\\
				\cline{2-5}
				\rule{0pt}{8pt}&Ours	&\textbf{51.14}	&\textbf{64.92}	&\textbf{82.54}\\
				\hline
			\end{tabular}
		}
	\end{center}
\end{table}


\begin{table}[t]
	\begin{center}
		\caption{Semantic segmentation results (\%) on the ScanNet dataset.} \label{tab_scannet}
		\resizebox{0.95\columnwidth}{!}{
			\begin{tabular}{|c|c|c|c|c|}
				\hline
				\rule{0pt}{8pt}&Methods	&mIoU	&mAcc	&OA\\
				\hline
				\multirow{7}{*}{\begin{sideways}20\%\end{sideways}}
				\rule{0pt}{8pt}&RandLA-Net~\cite{2020randlanet}	&52.86	&62.56	&81.43\\
				\rule{0pt}{8pt}&GA-Net~\cite{2021ganet}	&52.12	&61.50 	&81.39\\
				\rule{0pt}{8pt}&SCF-Net~\cite{2021scfnet}	&52.05	&61.32	&81.31\\			
				\cline{2-5}
				\rule{0pt}{8pt}&$\pi$-Model~\cite{2017pimodel}		&53.07	&62.78	&81.52\\
				\rule{0pt}{8pt}&Mean Teacher~\cite{2017meanteacher}	&52.98	&62.65	&81.48\\
				\rule{0pt}{8pt}&Pseudo-Labels~\cite{2013pseudolabels}	&53.23	&62.95	&81.63\\
				\cline{2-5}
				\rule{0pt}{8pt}&Ours	&\textbf{55.12}	&\textbf{63.61}	&\textbf{82.43}\\
				\hline
				\multirow{7}{*}{\begin{sideways}10\%\end{sideways}}
				\rule{0pt}{8pt}&RandLA-Net~\cite{2020randlanet}	&49.34	&58.20	&79.66\\
				\rule{0pt}{8pt}&GA-Net~\cite{2021ganet}	&49.03	&58.05	&79.29\\
				\rule{0pt}{8pt}&SCF-Net~\cite{2021scfnet}	&49.11	&59.35	&79.21\\			
				\cline{2-5}
				\rule{0pt}{8pt}&$\pi$-Model~\cite{2017pimodel}	&49.52	&58.48	&79.87\\
				\rule{0pt}{8pt}&Mean Teacher~\cite{2017meanteacher}	&49.41	&58.65	&79.70\\
				\rule{0pt}{8pt}&Pseudo-Labels~\cite{2013pseudolabels}	&50.25	&59.37	&79.92\\
				\cline{2-5}
				\rule{0pt}{8pt}&Ours	&\textbf{52.38}	&\textbf{60.76}	&\textbf{81.18}\\
				\hline
			\end{tabular}
		}
	\end{center}
\end{table}


\section{Experiments}
In this section, we firstly introduce the details of experimental setup. Secondly, we evaluate the performances of proposed superpoint-guided semi-supervised segmentation network on two 3D public datasets with a few labeled 3D scenes. Thirdly, we explore the effect of $t_{plo}$. Finally, we end up with ablation analysis.

\subsection{Experimental Setup}
The proposed superpoint-guided semi-supervised segmentation network is evaluated on two 3D public datasets, including S3DIS~\cite{2016s3dis}, and ScanNet~\cite{2017scannet}. 
In the geometry-based Region Growing algorithm~\cite{1994regiongrowin}, the curvature threshold value $t_{cvt}$ is $1$, and the angle threshold value $t_{ang}$ is $3$ degree following PCL~\cite{pclrg}. In the color-based Region Growing algorithm, the color threshold value $t_{clr}$ is $6$ following PCL~\cite{pclrgrgb}.
In the PLO module, the ratio parameter $t_{plo}$ is $0.8$.
The U-Net-based feature extractor parameters are consistent with the model before the FC layers in RandLA-Net~\cite{2020randlanet}, where $C_h$ is $64$. The output dimensionality of the first FC layer in EP module is $6$. We train the network using the Adam optimizer with initial learning rate $0.01$ and batchsize $6$ for $100$ epochs. In the first $50$ epochs, we only optimize the network branch for labeled point clouds. And in the last $50$ epochs, we train the whole network. The pseudo labels are updated after each epoch.

\subsection{Evaluation on the S3DIS Dataset}
The S3DIS dataset consists of 271 rooms in 6 different areas inside an office building. 13 semantic categories are assigned to each 3D point with XYZ coordinates and RGB features.
Since the fifth area with 68 rooms does not overlap with other areas, experiments on Area-5 could better reflect the generalization ability of the framework. So we conducted our experiments on Area-5 validation. 
We randomly sample about 20\% and 10\% (40 and 20 rooms) of the 203 rooms respectively in the training set as labeled point clouds, and the remaining rooms in the training set are used as unlabeled point clouds.
The evaluation metrics we use are mean class Intersection-over-Union (mIoU), mean class Accuracy (mAcc) and Overall Accuracy (OA).

We compare our superpoint-guided semi-supervised segmentation network to several state-of-the-art point cloud semantic segmentation methods with same labeled and unlabeled data including RandLA-Net~\cite{2020randlanet}, GA-Net~\cite{2021ganet}, and SCF-Net~\cite{2021scfnet}, and several popular semi-supervised semantic segmentation methods based on RandLA-Net including $\pi$-Model~\cite{2017pimodel}, Mean Teacher~\cite{2017meanteacher}, and Pseudo-Labels~\cite{2013pseudolabels}. In the $\pi$-Model and Mean Teacher, the dual inputs are the original point cloud and the point cloud after a random plane rotation and a random mirror transformation. In the Pseudo-Labels, the predicted labels are updated after each epoch. 
As seen from Table \ref{tab_s3dis}, 
$\pi$-Model and Mean Teacher only improving mIoU by about 1\% based on RandLA-Net indicates that the consistency between geometric transformed point clouds is not enough to constrain the unlabeled point cloud features. The results of Pseudo-Labels are worse than our method, indicating that there are some false-predicted pseudo labels which will affect the learning of network.
Our method achieves best on all metrics due to its more effective use of unlabeled data. 
The results on 20\% semi-supervised setting are better than on 10\% semi-supervised setting, which may be attributed to more labeled point clouds. 

\subsection{Evaluation on the ScanNet Dataset}
The ScanNet dataset contains 1,513 3D indoor scenes obtained by scanning and reconstruction, of which 1,201 are used for training and the remaining 312 are used for testing. 20 semantic categories are provided for evaluation. 
We randomly sample about 20\% and 10\% (240 and 120 rooms) of the 1201 rooms in the training set as labeled scenes, and the remaining rooms in the training set are used as unlabeled scenes.
The mIoU, mAcc, and OA are used as evaluation metrics. 

The competitive methods we use following experiments on the S3DIS dataset. Table \ref{tab_scannet} shows the comparison results. 
As seen from Table \ref{tab_scannet}, the results of $\pi$-Model~\cite{2017pimodel}, Mean Teacher~\cite{2017meanteacher}, and Pseudo-Labels~\cite{2013pseudolabels} have a small improvement on the basis of RandLA-Net, which may be attributed to the fact that there are more semantic categories in ScanNet than S3DIS, which results in a small number of labeled points of some categories. It is not easy to learn the features of these categories by the DNNs.
Our method achieves the state-of-the-art performance, probably due to the great pseudo-label filtering and feature constraining abilities.


\begin{table}[t]
	\begin{center}
		\caption{Results of different $t_{plo}$ on the S3DIS dataset (Area-5).} \label{tab_tplo}
		\resizebox{0.8\columnwidth}{!}{
			\begin{tabular}{|c|c|c|c|c|}
				\hline
				\rule{0pt}{8pt}&$t_{plo}$ values	&mIoU	&mAcc	&OA\\
				\hline
				\multirow{5}{*}{\begin{sideways}20\%\end{sideways}}
				\rule{0pt}{8pt}&0.70	&53.93	&64.80 	&82.92\\
				\rule{0pt}{8pt}&0.75	&54.48	&65.20	&83.23\\
				\rule{0pt}{8pt}&0.80	&\textbf{55.49}	&\textbf{65.45}	&\textbf{83.55}\\
				\rule{0pt}{8pt}&0.85	&54.13	&64.79	&82.81\\
				\rule{0pt}{8pt}&0.90	&53.38	&64.02	&82.53\\
				\hline
				\multirow{5}{*}{\begin{sideways}10\%\end{sideways}}
				\rule{0pt}{8pt}&0.70	&50.59	&64.56	&81.93\\
				\rule{0pt}{8pt}&0.75	&50.97	&64.78	&82.21\\
				\rule{0pt}{8pt}&0.80	&\textbf{51.14}	&\textbf{64.92}	&\textbf{82.54}\\
				\rule{0pt}{8pt}&0.85	&50.67	&63.56	&82.19\\
				\rule{0pt}{8pt}&0.90	&49.57	&60.07	&82.15\\
				\hline
			\end{tabular}
		}
	\end{center}
\end{table}


\begin{table}[t]
	\begin{center}
		\caption{Ablation study of the modules on the S3DIS dataset (Area-5).} \label{tab_module}
		\resizebox{1.0\columnwidth}{!}{
			\begin{tabular}{|c|c|c|c|c|}
				\hline
				\rule{0pt}{8pt}&Methods	&mIoU	&mAcc	&OA\\
				\hline
				\multirow{6}{*}{\begin{sideways}20\%\end{sideways}}
				\rule{0pt}{8pt}&Baseline	&50.90	&60.76	&81.24\\
				\rule{0pt}{8pt}&Baseline+SPFA	&51.32	&61.21	&82.22\\
				\rule{0pt}{8pt}&Baseline+SPFA+PL	&52.75	&63.86	&82.78\\			
				\rule{0pt}{8pt}&Baseline+SPFA+PLO	&53.95	&64.57	&83.04\\
				\rule{0pt}{8pt}&Baseline+SPFA+PLO+EP	&54.77	&64.98	&83.30\\
				\rule{0pt}{8pt}&Ours	&\textbf{55.49}	&\textbf{65.45}	&\textbf{83.55}\\
				\hline
				\multirow{6}{*}{\begin{sideways}10\%\end{sideways}}
				\rule{0pt}{8pt}&Baseline	&45.64	&58.58	&79.08\\
				\rule{0pt}{8pt}&Baseline+SPFA	&46.05	&59.62	&80.38\\
				\rule{0pt}{8pt}&Baseline+SPFA+PL	&47.86	&61.59	&81.38\\			
				\rule{0pt}{8pt}&Baseline+SPFA+PLO	&49.78	&62.63	&81.86\\
				\rule{0pt}{8pt}&Baseline+SPFA+PLO+EP	&50.45	&63.25	&82.17\\
				\rule{0pt}{8pt}&Ours	&\textbf{51.14}	&\textbf{64.92}	&\textbf{82.54}\\
				\hline
			\end{tabular}
		}
	\end{center}
\end{table}

\subsection{Effect of $t_{plo}$}
The ratio parameter $t_{plo}$ in the PLO module affects the quality of the optimized pseudo labels, and results in affecting the final segmentation performances. 
Too small value of $t_{plo}$ will result in pseudo labels with lower-confidence being assigned to superpoints, and too large value of $t_{plo}$ will result in many correct pseudo labels being deleted.
Here we conduct experiments to analyze the effect of $t_{plo}$ by setting different values \{0.7, 0.75, 0.8, 0.85, 0.9\}.
We conduct experiments on Area-5 of the S3DIS dataset with the evaluation metrics mIoU, mAcc and OA. 
The results are listed in Table \ref{tab_tplo}. As seen from Table \ref{tab_tplo}, the results by setting $t_{plo}$ to 0.8 achieve the
best performance, so we use this value as $t_{plo}$ in the PLO module.

\subsection{Ablation Study}
For ablation study, we stack the proposed sub-modules on the baseline step-to-step to prove the effectiveness of our method.
Our baseline method employs a U-Net-based feature extractor from RandLA-Net~\cite{2020randlanet}, and is only trained on the labeled point clouds for $100$ epochs. 
The comparing experiments are (1) baseline method, denoted as ``Baseline"; (2) adding the SPG and SPFA modules on baseline and being trained on the labeled point clouds, denoted as ``Baseline+SPFA"; (3) adding pseudo labels to unlabeled point clouds for supervision in the last $50$ epochs based on (2), denoted as ``Baseline+SPFA+PL"; (4) adding the PLO module on (3) for unlabeled point clouds, denoted as ``Baseline+SPFA+PLO"; (5) adding the EP module on (4) for all point clouds, denoted as ``Baseline+SPFA+PLO+EP"; and (6) adding the superpoint feature consistency loss functions $Loss^l_{sp}$ and $Loss^u_{sp}$ on (5), denoted as ``Ours". 
We conduct ablation study on Area-5 of the S3DIS dataset with the evaluation metrics mIoU, mAcc and OA. And 20\% and 10\% of the rooms in the training set are used for labeled point clouds.

As shown in Table \ref{tab_module}, the performances on ``Baseline+SPFA" being better than ``Baseline" demonstrate the importance of smoothing the features in superpoints. ``Baseline+SPFA+PL" achieves better than ``Baseline+SPFA", which may be attributed to the supervision of unlabeled point clouds. ``Baseline+SPFA+PLO" performing better than ``Baseline+SPFA+PL" indicates that the superpoints produced by combining geometry-based and color-based Region Growing algorithms~\cite{1994regiongrowin} can help optimize pseudo labels effectively. The result of ``Baseline+SPFA+PLO+EP" achieves better than ``Baseline+SPFA+PLO", which may be attributed to edge-point feature learning. ``Ours" achieves best, which demonstrates that combining all these modules can reach the best results.

\section{Conclusions}
For using the large amount of unlabeled point clouds which can be easily obtained from sensors or reconstruction, we propose a superpoint-guided semi-supervised segmentation network for 3D point clouds, which jointly utilizes a small portion of labeled scenes and a large number of unlabeled scenes for network training. 
Specifically, we combine the superpoints produced by geometry-based and color-based Region Growing algorithms~\cite{1994regiongrowin} to optimize the pseudo labels predicted by unlabeled point clouds. 
The features of points without pseudo labels are constrained by the superpoint feature aggregation module, the edge prediction module, and the superpoint feature consistency loss function. 
Our method can learn the discriminative features of unlabeled point clouds and achieve best performance on two 3D public datasets with a few number of labeled scenes in most cases.



\newpage
\bibliographystyle{IEEEtran}
\bibliography{root}

\begin{thebibliography}{10}
\providecommand{\url}[1]{#1}
\csname url@rmstyle\endcsname
\providecommand{\newblock}{\relax}
\providecommand{\bibinfo}[2]{#2}
\providecommand\BIBentrySTDinterwordspacing{\spaceskip=0pt\relax}
\providecommand\BIBentryALTinterwordstretchfactor{4}
\providecommand\BIBentryALTinterwordspacing{\spaceskip=\fontdimen2\font plus
\BIBentryALTinterwordstretchfactor\fontdimen3\font minus
  \fontdimen4\font\relax}
\providecommand\BIBforeignlanguage[2]{{%
\expandafter\ifx\csname l@#1\endcsname\relax
\typeout{** WARNING: IEEEtran.bst: No hyphenation pattern has been}%
\typeout{** loaded for the language `#1'. Using the pattern for}%
\typeout{** the default language instead.}%
\else
\language=\csname l@#1\endcsname
\fi
#2}}

\bibitem{2017pointnet}
C.~R. Qi, H.~Su, K.~Mo, and L.~J. Guibas, ``Pointnet: Deep learning on point
  sets for 3d classification and segmentation,'' in \emph{Proceedings of the
  IEEE Conference on Computer Vision and Pattern Recognition (CVPR)}, 2017, pp.
  652--660.

\bibitem{2017pointnet++}
C.~R. Qi, L.~Yi, H.~Su, and L.~J. Guibas, ``Pointnet++: Deep hierarchical
  feature learning on point sets in a metric space,'' in \emph{Proceedings of
  the Conference on Neural Information Processing Systems (NeurIPS)}, 2017, pp.
  5099--5108.

\bibitem{2018sscn}
B.~Graham, M.~Engelcke, and L.~van~der Maaten, ``3d semantic segmentation with
  submanifold sparse convolutional networks,'' in \emph{Proceedings of the IEEE
  Conference on Computer Vision and Pattern Recognition (CVPR)}, 2018, pp.
  9224--9232.

\bibitem{2018pointcnn}
Y.~Li, R.~Bu, M.~Sun, W.~Wu, X.~Di, and B.~Chen, ``Pointcnn: Convolution on
  x-transformed points,'' in \emph{Proceedings of the Conference on Neural
  Information Processing Systems (NeurIPS)}, 2018, pp. 820--830.

\bibitem{2018spgraph}
L.~Landrieu and M.~Simonovsky, ``Large-scale point cloud semantic segmentation
  with superpoint graphs,'' in \emph{Proceedings of the IEEE Conference on
  Computer Vision and Pattern Recognition (CVPR)}, 2018, pp. 4558--4567.

\bibitem{2018dgcnn}
Y.~Wang, Y.~Sun, Z.~Liu, S.~E. Sarma, M.~M. Bronstein, and J.~M. Solomon,
  ``Dynamic graph cnn for learning on point clouds,'' \emph{ACM Transaction on
  Graphics (TOG)}, vol.~38, pp. 1--12, 2019.

\bibitem{2019pointweb}
H.~Zhao, L.~Jiang, C.-W. Fu, and J.~Jia, ``Pointweb: Enhancing local
  neighborhood features for point cloud processing,'' in \emph{Proceedings of
  the IEEE Conference on Computer Vision and Pattern Recognition (CVPR)}, 2019,
  pp. 5565--5573.

\bibitem{2019sspspg}
L.~Landrieu and M.~Boussaha, ``Point cloud oversegmentation with
  graph-structured deep metric learning,'' in \emph{Proceedings of the IEEE
  Conference on Computer Vision and Pattern Recognition (CVPR)}, 2019, pp.
  7432--7441.

\bibitem{2020randlanet}
Q.~Hu, B.~Yang, L.~Xie, S.~Rosa, Y.~Guo, Z.~Wang, N.~Trigoni, and A.~Markham,
  ``Randla-net: Efficient semantic segmentation of large-scale point clouds,''
  in \emph{Proceedings of the IEEE Conference on Computer Vision and Pattern
  Recognition (CVPR)}, 2020, pp. 11\,105--11\,114.

\bibitem{2020grid}
Q.~Xu, X.~Sun, C.-Y. Wu, P.~Wang, and U.~Neumann, ``Grid-gcn for fast and
  scalable point cloud learning,'' in \emph{Proceedings of the IEEE Conference
  on Computer Vision and Pattern Recognition (CVPR)}, 2020, pp. 5661--5670.

\bibitem{2021rtn}
S.~Deng, B.~Liu, Q.~Dong, and Z.~Hu, ``Rotation transformation network:
  Learning view-invariant point cloud for classification and segmentation,'' in
  \emph{Proceedings of the IEEE International Conference on Multimedia and Expo
  (ICME)}, 2021, pp. 1--6.

\bibitem{2021scfnet}
S.~Fan, Q.~Dong, F.~Zhu, Y.~Lv, P.~Ye, and F.-Y. Wang, ``Scf-net: Learning
  spatial contextual features for large-scale point cloud segmentation,'' in
  \emph{Proceedings of the IEEE Conference on Computer Vision and Pattern
  Recognition (CVPR)}, 2021, pp. 14\,504--14\,513.

\bibitem{2021ganet}
S.~Deng and Q.~Dong, ``Ga-net: Global attention network for point cloud
  semantic segmentation,'' \emph{IEEE Signal Processing Letters (SPL)},
  vol.~28, pp. 1300--1304, 2021.

\bibitem{2012active}
Y.~Wang, S.~Asafi, O.~van Kaick, H.~Zhang, D.~Cohen-Or, and B.~Chen, ``Active
  co-analysis of a set of shapes,'' \emph{ACM Transactions on Graphics (TOG)},
  vol.~31, pp. 1--10, 2012.

\bibitem{2020wsspcs}
X.~Xu and G.~H. Lee, ``Weakly supervised semantic point cloud segmentation:
  Towards 10x fewer labels,'' in \emph{Proceedings of the IEEE Conference on
  Computer Vision and Pattern Recognition (CVPR)}, 2020, pp. 13\,706--13\,715.

\bibitem{2020ss3ld}
J.~Mei, B.~Gao, D.~Xu, W.~Yao, X.~Zhao, and H.~Zhao, ``Semantic segmentation of
  3d lidar data in dynamic scene using semi-supervised learning,'' \emph{IEEE
  Transactions on Intelligent Transportation Systems (TITS)}, vol.~21, pp.
  2496--2509, 2020.

\bibitem{2021spcs}
H.~Li, Z.~Sun, Y.~Wu, and Y.~Song, ``Semi-supervised point cloud segmentation
  using self-training with label confidence prediction,''
  \emph{Neurocomputing}, vol. 437, pp. 227--237, 2021.

\bibitem{2021sspcnet}
M.~Cheng, L.~Hui, J.~Xie, and J.~Yang, ``Sspc-net: Semi-supervised semantic 3d
  point cloud segmentation network,'' in \emph{Proceedings of the AAAI
  Conference on Artificial Intelligence (AAAI)}, 2021, pp. 1140--1147.

\bibitem{2017pimodel}
S.~Laine and T.~Aila, ``Temporal ensembling for semi-supervised learning,'' in
  \emph{Proceedings of the International Conference on Learning Representations
  (ICLR)}, 2017, pp. 1--13.

\bibitem{2017meanteacher}
A.~Tarvainen and H.~Valpola, ``Mean teachers are better role models:
  Weight-averaged consistency targets improve semi-supervised deep learning
  results,'' in \emph{Proceedings of the Conference on Neural Information
  Processing Systems (NeurIPS)}, 2017, p. 1195–1204.

\bibitem{2017ssgan}
N.~Souly, C.~Spampinato, and M.~Shah, ``Semi supervised semantic segmentation
  using generative adversarial network,'' in \emph{Proceedings of the IEEE
  International Conference on Computer Vision (ICCV)}, 2017, pp. 5689--5697.

\bibitem{2021dtc}
X.~Luo, J.~Chen, T.~Song, and G.~Wang, ``Semi-supervised medical image
  segmentation through dual-task consistency,'' in \emph{Proceedings of the
  AAAI Conference on Artificial Intelligence (AAAI)}, 2021, pp. 8801--8809.

\bibitem{2020pointnl}
M.~Cheng, L.~Hui, J.~Xie, J.~Yang, and H.~Kong, ``Cascaded non-local neural
  network for point cloud semantic segmentation,'' in \emph{Proceedings of the
  IEEE/RSJ International Conference on Intelligent Robots and Systems (IROS)},
  2020, pp. 8447--8452.

\bibitem{1994regiongrowin}
R.~Adams and L.~Bischof, ``Seeded region growing,'' \emph{IEEE Transactions on
  Pattern Analysis and Machine Intelligence (TPAMI)}, vol.~16, pp. 641--647,
  1994.

\bibitem{2015unet}
O.~Ronneberger, P.~Fischer, and T.~Brox, ``U-net: Convolutional networks for
  biomedical image segmentation,'' in \emph{International Conference on Medical
  Image Computing and Computer-Assisted Intervention (MMICCAI)}, 2015, pp.
  234--241.

\bibitem{2013lrelu}
A.~L. Maas, A.~Y. Hannun, and A.~Y. Ng, ``Rectifier nonlinearities improve
  neural network acoustic models,'' in \emph{Proceedings of the International
  Conference on Machine Learning (ICML)}, 2013, pp. 1--6.

\bibitem{2013pseudolabels}
D.-H. Lee, ``Pseudo-label : The simple and efficient semi-supervised learning
  method for deep neural networks,'' in \emph{Proceedings of the International
  Conference on Machine Learning Workshop (ICMLW)}, 2013, pp. 896--901.

\bibitem{2016s3dis}
I.~Armeni, O.~Sener, A.~R. Zamir, H.~Jiang, I.~Brilakis, M.~Fischer, and
  S.~Savarese, ``3d semantic parsing of large-scale indoor spaces,'' in
  \emph{Proceedings of the IEEE Conference on Computer Vision and Pattern
  Recognition (CVPR)}, 2016, pp. 1534--1543.

\bibitem{2017scannet}
A.~Dai, A.~X. Chang, M.~Savva, M.~Halber, T.~Funkhouser, and M.~Nie{\ss}ner,
  ``Scannet: Richly-annotated 3d reconstructions of indoor scenes,'' in
  \emph{Proceedings of the IEEE Conference on Computer Vision and Pattern
  Recognition (CVPR)}, 2017, pp. 2432--2443.

\bibitem{pclrg}
S.~Ushakov, ``Region growing segmentation,''
  \url{https://pcl.readthedocs.io/projects/tutorials/en/latest/region_growing_segmentation.html?highlight=region%20growing}.

\bibitem{pclrgrgb}
------, ``Color-based region growing segmentation,''
  \url{https://pcl.readthedocs.io/projects/tutorials/en/latest/region_growing_rgb_segmentation.html?highlight=region%20growing}.

\end{thebibliography}

\end{document}